\documentclass[twocolumn,letterpaper]{IEEEAerospaceCLS}  

\usepackage[]{graphicx, amssymb, amsmath}

\newcommand{\ignore}[1]{}  

\begin{document}
\title{Enhancing Rover Mobility Monitoring: Autoencoder-driven Anomaly Detection for Curiosity}

\author{%
Mielad Sabzehi\\ 
Jet Propulsion Laboratory\\
California Institute of Technology\\
4800 Oak Grove Dr.\\
Pasadena, CA 91109\\
mielad.sabzehi@jpl.nasa.gov
\and 
Peter Rollins\\ 
Jet Propulsion Laboratory\\
California Institute of Technology\\
4800 Oak Grove Dr.\\
Pasadena, CA 91109\\
peter.j.rollins@jpl.nasa.gov
\thanks{\footnotesize 979-8-3503-0462-6/24/$\$31.00$ }              
\thanks{{\copyright2024 IEEE.  Personal use of this material is permitted.  Permission from IEEE must be obtained for all other uses, in any current or future media, including reprinting/republishing this material for advertising or promotional purposes, creating new collective works, for resale or redistribution to servers or lists, or reuse of any copyrighted component of this work in other works.}}         
}

\maketitle

\thispagestyle{plain}
\pagestyle{plain}

\maketitle

\thispagestyle{plain}
\pagestyle{plain}

\begin{abstract}
Over eleven years into its mission, the Mars Science Laboratory remains vital to NASA's Mars exploration. Safeguarding the rover's long-term functionality is a top mission priority. In this study, we introduce and test undercomplete autoencoder models for detecting drive anomalies, using telemetry data from wheel actuators, the Rover Inertial Measurement Unit (RIMU), and the suspension system. Our approach enhances post-drive data analysis during tactical downlink sessions. We explore various model architectures and input features to understand their impact on performance. Evaluating the models involves testing them on unseen data to mimic real-world scenarios. Our experiments demonstrate the undercomplete autoencoder model's effectiveness in detecting drive anomalies within the Curiosity rover dataset. Remarkably, the model even identifies subtle anomalous telemetry patterns missed by human operators. Additionally, we provide insights into optimal design choices by comparing different model architectures and input features. The model's ability to capture inconspicuous anomalies, potentially indicating early-stage failures, holds promise for the field, by improving the reliability and safety of future planetary exploration missions through early anomaly detection and proactive maintenance.
\end{abstract}

\tableofcontents

\section{Introduction}
The Mars Science Laboratory (MSL), a pivotal asset in NASA's Mars exploration program, has significantly contributed to our understanding of the Martian environment over the past eleven years. Safeguarding the operational integrity of the rover is paramount to ensure its continued functionality for years to come. During tactical operations \cite{Gildner.2021}, a crucial aspect is the rapid assessment of rover health by the engineering team through flight telemetry data received during downlink. The downlink operator meticulously evaluates this engineering telemetry to gauge the rover's state, health, and overall performance. This assessment encompasses the success of the prior sol’s activity plan execution, critical subsystem statuses, power management, and instrument health. A notable aspect of this mission involves developing strategies for effective rover monitoring and anomaly detection to proactively address potential technical issues. An autonomous model screening telemetry for anomalies could significantly aid human operators in not overlooking potential issues.

Previous research by Sabzehi et al. emphasized the potential of undercomplete autoencoders for feature-learning, particularly for estimating proprioceptive slip in planetary exploration rovers \cite{Sabzehi.2023}. Building upon this approach, we present the development and testing of undercomplete encoder-decoder neural network models, commonly known as autoencoders, designed for drive anomaly detection. These models utilize telemetry data sourced from wheel actuators, the Inertial Measurement Unit (IMU), and the suspension system.

Our approach introduces CAIDDA (\textbf{C}uriosity \textbf{AI} \textbf{D}etects \textbf{D}rive \textbf{A}nomalies), a robust tool for post-drive ground data analysis during tactical downlink sessions. CAIDDA utilizes undercomplete autoencoders, trained to compress (encode) input data into a low-dimensional latent space and then accurately reconstruct (decode) the information back to its original dimension. The dimensionality reduction prompts the model to prioritize the most critical features. Anomalies, or deviations from the learned patterns, are likely to result in higher reconstruction errors when the autoencoder attempts to reproduce the input data. By quantifying these errors, drive telemetry exhibiting high error rates can be flagged, facilitating efficient anomaly detection. The models are trained using mission data collected over 3815 Martian days (Sols).

This work delves into a comprehensive exploration of different input features, analyzing their impact on model performance. The model's performance is evaluated by applying the trained autoencoder to unseen and untrained data, mimicking real-world scenarios. Our experimental results effectively showcase the undercomplete autoencoder model's prowess in detecting drive anomalies within the Curiosity rover dataset. Remarkably, the model even highlighted subtle anomalous telemetry patterns that human operators were unable to detect. Additionally, by comparing different model architectures and input features, we provide valuable insights into optimal design choices for this specific anomaly detection task.

The research outlined in this work significantly contributes to the field of anomaly detection for space flight operations. The developed approach holds promise in enhancing the reliability and safety of future planetary exploration missions by enabling early anomaly detection and facilitating proactive maintenance interventions. Furthermore, the model's capacity to discern intricate patterns from complex telemetry data makes it optimal for capturing subtle anomalies that may serve as early indicators of potential failures.

We begin by presenting an overview of relevant literature on this topic. In Chapter 3, we detail our methodology, starting with an introduction to the mission data utilized, followed by our preprocessing and feature engineering approach, the training of the autoencoder, and its application for anomaly detection. Moving forward, Chapter 4 entails an experimental evaluation, simulating a tactical deployment of CAIDDA. We conclude with a concise summary and an outlook on future directions in the final section.

\section{Related Work}
Anomaly detection is a critical field of research in various domains, including space exploration, finance, cybersecurity, and more \cite{LAKHMIRI2022116060}. It pertains to the identification of unusual patterns that do not conform to expected behavior. These outliers can often signify issues such as system faults and mechanical defects. The role of anomaly detection in space flight missions like the Mars Science Laboratory is of significant importance in ensuring operational integrity and mission success.

The analysis of telemetry data from Mars is an operation characterized by time sensitivity. Engineering teams operate under significant time constraints to evaluate the volume of data transmitted during downlink sessions.  The rapid assessment and decision-making process would benefit from robust and efficient tools capable of automatically identifying anomalies that may otherwise be overlooked in manual analysis \cite{Gildner.2021}.

Classical Model-based fault detection models require a deep understanding of the system and the environment. Contrary to that, data-driven models excel at identifying subtle, unanticipated anomalies \cite{Chandola.2009}.

Lakhmiri et al. address the critical task of monitoring telemetry data sent by the Mars Science Laboratory rover. This telemetry data, prone to corruption and volume loss due to various factors such as cosmic events and planetary movements, requires prompt and accurate analysis to ensure its integrity. The authors approach this problem as an anomaly detection task, emphasizing the importance of quick and precise differentiation between complete and incomplete passes of data transmission. They spotlight the significant role of automation in enhancing this process, particularly in light of the deployment of team members to new missions. The study extensively employs Variational Autoencoders (VAEs), a form of deep neural networks, optimized for anomaly detection in this specific context. The presented approach lead to high accuracy in identifying anomalies in the telemetry data, ensuring the timely and reliable transmission of information from the Mars Curiosity rover \cite{LAKHMIRI2022116060}.

Autoencoders \cite{LeCun.1989} \cite{Bourlard.1988} are a type of neural network utilized for data-driven anomaly detection. Comprising an encoder $h = e(x)$ and a decoder $\tilde{x} = d(h)$, they are trained to output $\tilde{x}$ as a reflection of the input $x$. A successful autoencoder aims for $\tilde{x} = x$, but constraints in undercomplete autoencoders avert mere input duplication. These constraints necessitate $h$ to possess a smaller dimension than $x$, compelling the encoder to condense the data and the decoder to rebuild it from this condensed form, creating a 'bottleneck'. This configuration assists in learning crucial data representations, analogous to PCA with linear decoders and mean squared error loss. Employing nonlinear functions in encoders and decoders permits advanced learning beyond PCA \cite{Goodfellow.2016}. Anomaly detection with autoencoders is accomplished by examining the reconstruction error between $x$ and $\tilde{x}$. Elevated errors indicate anomalies, as they reveal difficulties in decoding the input data, allowing for efficient anomaly detection with a predefined error threshold.

\section{Methodology}\label{methodology}
In this chapter, we unveil the essential steps guiding our methodology in improving the detection of drive anomalies vital for the Mars Science Laboratory mission. We explain how we gathered and prepared the data, crafted an autoencoder tailored to our needs, and employed it to spot unusual patterns within the telemetry data.
\subsection{Data}
\begin{figure}[h]
\centering
\includegraphics[width=3in]{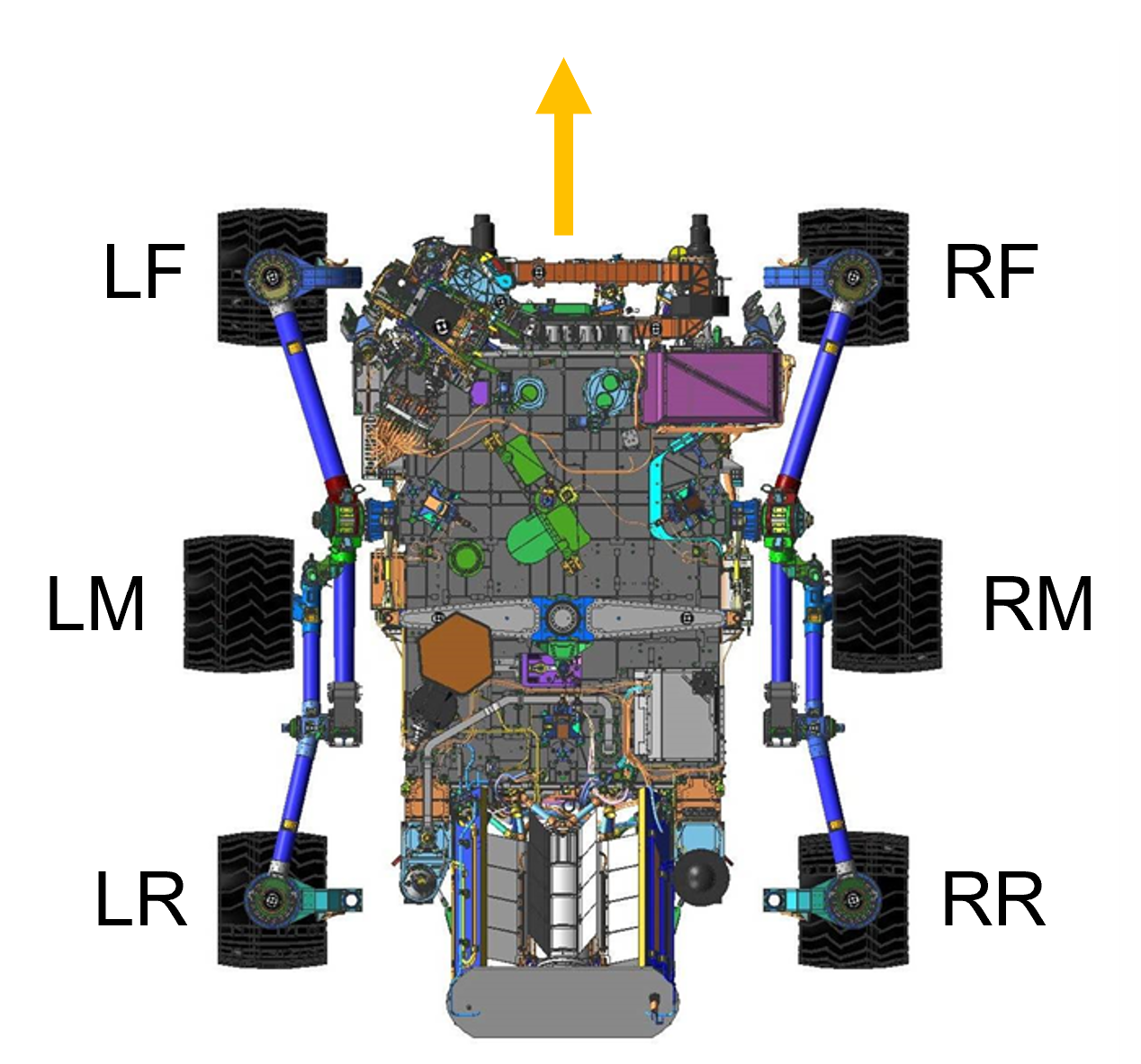}
\caption{\bf{Overview of Rover Wheel Annotation}}
\label{rover_wheel_connotation}
\end{figure}

With an odometer\footnote{As of September 2023} of ~31,302 m Curiosity offers an extensive database of mobility telemetry spreading across various types of martian terrain, potentially enabling the proposed data-driven mobility anomaly detection approach. The 8 Hz mobility telemetry used in this work is provided by the following rover components: six drive actuators, the Rover Inertia Measurement Unit (RIMU), two bogie resolvers (left and right), and two rocker resolvers (left and right). Figure \ref{rover_wheel_connotation} offers an overview of the annotation for the six drive actuators that we will refer to. The drive actuators provide current estimates, angular rates, and voltages. The RIMU provides accelerations and angular rates in and around all three Cartesian directions. Finally, the bogie and rocker resolvers provide their respective suspension angles. Using the available mobility telemetry signals, we indirectly calculate a set of additional signals as described in the next section. Table \ref{telemetry_signals} offers an overview of the available and indirectly calculated telemetry signals. 
To simplify the problem and reduce the generalization requirements for CAIDDA, we filter the mobility data for straight drives. Future versions of CAIDDA may be expanded to drives that include arcs and turn-in-place. 

\subsection{Preprocessing and Feature Engineering}
As described in the previous section, a set signals is calculated indirectly. Using the controlled voltage $U_{k}$ and current estimate $I_{k}$ of each wheel $k$, the power $P_{k}$ is calculated:
\begin{equation}
    P_{k} = I_{k} \cdot U_{k}
\end{equation}
In order to obtain a metric that expresses the inequality of current demand between wheels, Ugenti et al. \cite{Ugenti.2022} proposes the computation of the current deviation. The current deviation, $CD$, of wheel $k$ is defined as:
\begin{equation}
    CD(I_{k}) = \lvert I_{k}-\Bar{I} \rvert
\end{equation}
Where $I_{k}$ represents the current of wheel $k$ and $\Bar{I}$ the mean over  all $N=6$ wheel currents. $\Bar{I}$ may be calculated with:
\begin{equation}
    \Bar{I}=\frac{\sum_{k=1}^{N=6} I_{k}}{N}
\end{equation}
This indirectly calculated signal provides a measure of unequal current demand between wheels. $CD$ becomes large for wheels with a relatively high current draw and small for wheels with a relatively low current draw. In a similar manner, the power deviation $PD$ for each wheel is calculated:
\begin{equation}
    PD(P_{k}) = \lvert P_{k}-\Bar{P} \rvert
\end{equation}
And $\Bar{P}$ being the mean over  all $N=6$ wheel powers:
\begin{equation}
    \Bar{P}=\frac{\sum_{k=1}^{N=6} P_{k}}{N}
\end{equation}
A total of 28 signals are provided from the sensors. Additionally, 18 signals are computed indirectly leading to a total of 46 signals as depicted in Table \ref{telemetry_signals}.
\begin{table}[h]
\renewcommand{\arraystretch}{1.3}
\caption{\bf Mobility Telemetry Signals}
\label{telemetry_signals}
\centering
\begin{tabular}{|c|l|l|}
\hline
\multicolumn{1}{|c|}{\textbf{Sensor}} & \textbf{Signal}   & \textbf{Source} \\ \hline
Drive Actuators             & Current (C)           & Sensor                \\ \cline{2-3} 
      (6x)                                & Current Deviation (CD) & Computed              \\ \cline{2-3} 
                                      & Angular Rate (Rate)      & Sensor                \\ \cline{2-3} 
                                      & Voltage (U)           & Sensor                \\ \cline{2-3} 
                                      & Power (P)             & Computed              \\ \cline{2-3} 
                                      & Power Deviation (PD)   & Computed              \\ \hline
RIMU                 & Acceleration X (Accel X)    & Sensor                \\ \cline{2-3} 
                                      & Acceleration Y (Accel Y)    & Sensor                \\ \cline{2-3} 
                                      & Acceleration Z (Accel Z)    & Sensor                \\ \cline{2-3} 
                                      & Rotation X (RX)        & Sensor                \\ \cline{2-3} 
                                      & Rotation Y (RY)        & Sensor                \\ \cline{2-3} 
                                      & Rotation Z RZ)        & Sensor                \\ \hline
Bogie                & Angle Left (BogieLeft)        & Sensor                \\ \cline{2-3} 
                                      & Angle Right (BogieRight)       & Sensor                \\ \hline
Differential         & Angle Left (DiffLeft)        & Sensor                \\ \cline{2-3} 
                                      & Angle Right (DiffRight)       & Sensor                \\ \hline
\end{tabular}
\end{table}

\begin{figure}[h]
\centering
\includegraphics[width=3in]{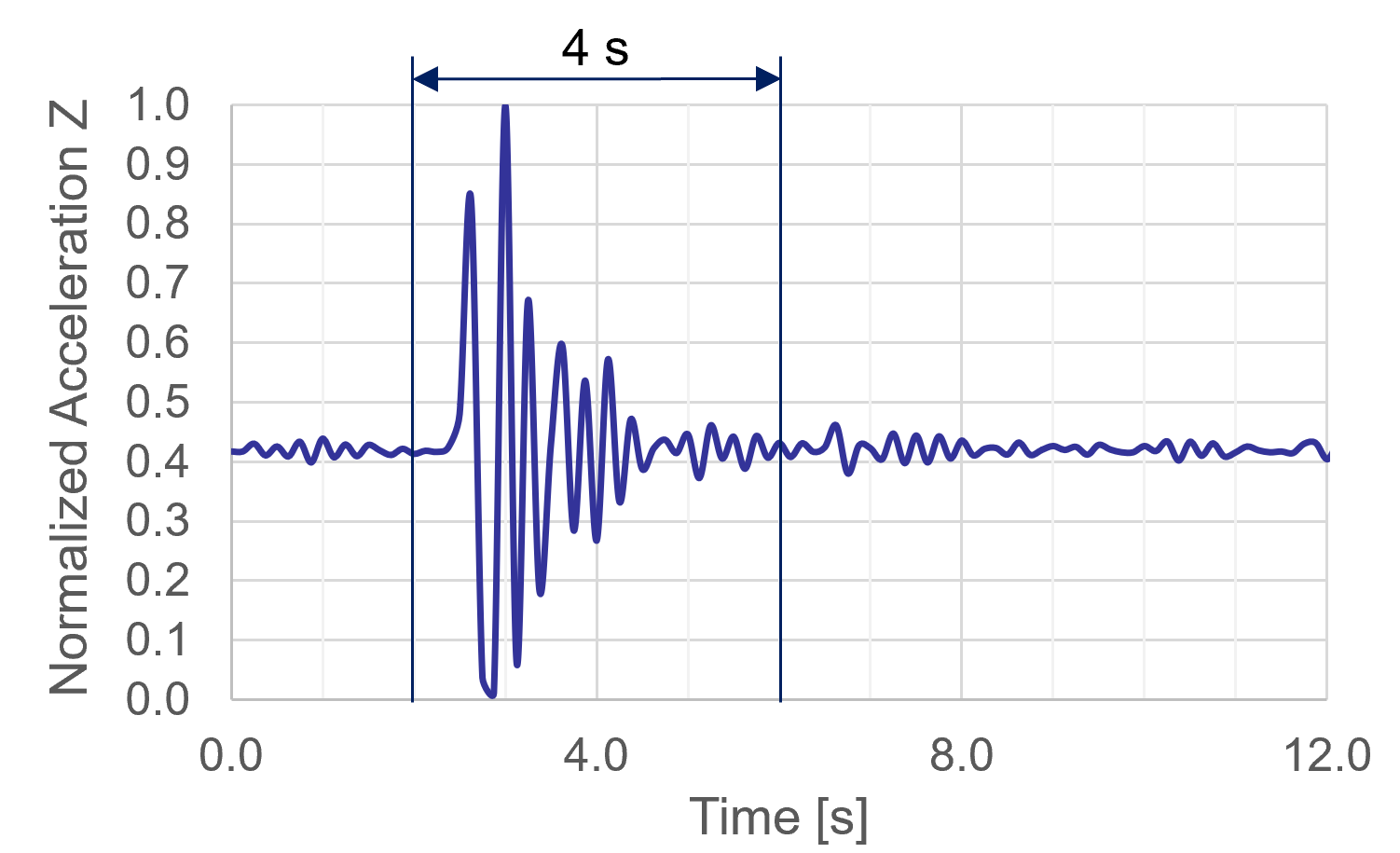}
\caption{\bf{Acceleration during Rock Traversal}}
\label{window_size}
\end{figure}

The time series mobility signals are split into samples by applying a 4 second rolling window with a stride of 1 second. The window length and stride are technically tunable parameters that could be optimized. The 4-second window size for the rolling window sampling was determined based on observations of rover acceleration during encounters with significant rocky terrain. The telemetry data showed that the drop and suspension oscillations usually stabilize within this timeframe after traversing larger rocks. Figure \ref{window_size} is illustrating this behavior for reference. Future work may evolve around a comparative variation of these parameters and their influence on anomaly detection capabilities. The time series sampling results in a total of 521,673 samples spanning 3815 martian days.

The time series samples are then transformed from a time-space representation to a feature-space representation by extracting seven key statistical features: mean, standard deviation (std), kurtosis (kurt), skewness (skew), minimum (min), maximum (max), and median. The transformation results in a total of 322 scalar features.
Since the feature values vary wildly in dimension, a scaling of the features is performed to improve the model performance \cite{Ioffe.2015}. The features are rescaled with a min-max normalization to range between 0 and 1 by determining the minimum and maximum value of each feature over all available samples.

\subsection{Autoencoder Development}
This section shall cover the development and training of the proposed undercomplete encoder-decoder neural network. Two different architectures are trained, which differ in the amount of input features. The base version of CAIDDA, CAIDDA-Prime, uses all 322 features explained in the previous section, whereas CAIDDA-Refined does not include acceleration-based features leading to 301 input features. This distinction in input features and models is chosen since early testing of CAIDDA-Prime indicated a strong bias of the anomaly detection towards acceleration based input features. To allow the detection of acceleration-independent anomalies a reduced input feature space is introduced with CAIDDA-Refined. 

\begin{figure*}[]
\centering
\includegraphics[width=6in]{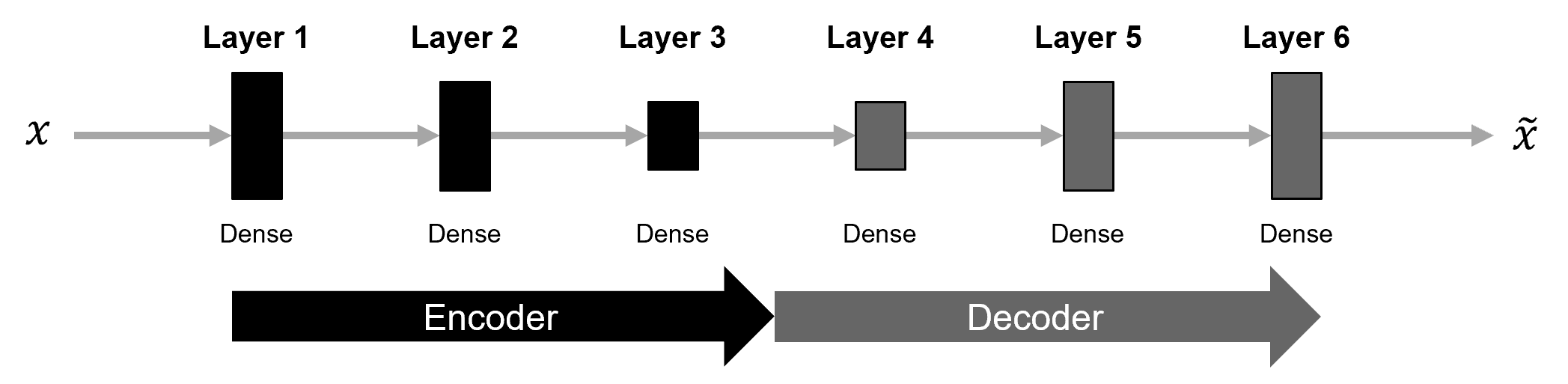}
\caption{\bf{Proposed Autoencoder Architecture}}
\label{aec_architecture_overview}
\end{figure*}

Figure \ref{aec_architecture_overview} offers a overview of the proposed autoencoder architecture. The encoder and decoder each consist of three dense (feedforward) layers. Table \ref{aec_architecture_detail} provides details of the actual neural network architecture by providing number of neurons and activation functions per layer for each model. Since the proposed autoencoder is symmetrical by design choice the repeating parameters for layer 4 to 6 are not provided. The proposed network parameters are determined by manually testing different architectures. Initially, our plan involved employing hyperparameter sweeps with wandb \cite{Biewald.2020} and utilizing Bayesian optimization to search for optimal network architectures. However, due to time constraints, we chose not to pursue this approach further.

\begin{table}[h]
\renewcommand{\arraystretch}{1.3}
\caption{\bf Detailed Architecture Parameters}
\label{aec_architecture_detail}
\centering
\begin{tabular}{|c|c|c|c|}
\hline
\multicolumn{1}{|c|}{\textbf{Model}}                                           & \multicolumn{1}{l|}{\textbf{Layer}} & \multicolumn{1}{l|}{\textbf{Neurons}} & \multicolumn{1}{l|}{\textbf{Activation}} \\ \hline
\textbf{}                                                                      & 1, 6                                   & 322                                   & linear                                   \\ \cline{2-4} 
\textbf{\begin{tabular}[c]{@{}c@{}}CAIDDA-Prime\end{tabular}}    & 2, 5                                   & 182                                   & sigmoid                                  \\ \cline{2-4} 
\textbf{}                                                                      & 3, 4                                   & 143                                   & linear                                   \\ \hline
\textbf{}                                                                      & 1, 6                                   & 301                                   & linear                                   \\ \cline{2-4} 
\textbf{\begin{tabular}[c]{@{}c@{}}CAIDDA-Refined\end{tabular}} & 2, 5                                   & 176                                   & sigmoid                                  \\ \cline{2-4} 
\textbf{}                                                                      & 3, 4                                   & 141                                   & linear                                   \\ \hline
\end{tabular}
\end{table}
Both models are trained with the Python library Keras \cite{Chollet.2015}, using the ADAM optimizer as the gradient descent algorithm. Each model is trained with 200 epochs using a mean-squared-error loss function. 20 \% of the samples are randomly selected as validation data to ensure a converging training process and the absence of overfitting. The examination of the loss plot affirms a convergence of training and validation loss for both models. There is no discernible deviation between training and validation loss, suggesting an absence of overfitting.

\subsection{Anomaly Detection}
In the preceding section, we elaborated on the development of the autoencoder models. In this section, we transition to the practical implementation of anomaly detection. This involves quantifying the reconstruction error of the autoencoders and leveraging this metric to identify and flag anomalies within the data. 

The reconstruction error vector $\mathbf{e}$ is the offset between model input $\mathbf{x}$ and model output $\mathbf{\tilde{x}}$:
\begin{equation}
    \mathbf{e}= \mathbf{x} - \mathbf{\tilde{x}}
\end{equation}
Each elements of vector $\mathbf{e}$ quantifies how well the autoencoder manages to reconstruct each feature from the latent space representation. To generate a single metric $a$ that quantifies the reconstruction error for the whole error vector we compute the 1-norm of $\mathbf{e}$:
\begin{equation}
    a = \Vert\mathbf{e}\Vert_{1}
\end{equation}

Using the reconstruction error $a$, a sample may be flagged as anomalous if $a$ exceeds a certain threshold. The threshold is determined by calculating the 99.9th percentile of the reconstruction error distribution observed in the training data. This percentile computation serves as the basis for establishing the cutoff value, set at the last value within the determined percentile. The 99.9th percentile threshold was chosen to streamline model validation and evaluation by minimizing the samples requiring in-depth investigation. This prioritizes precise identification of significant anomalies. However, for a more sensitive approach encompassing a broader range of anomalies, the percentile threshold can be lowered accordingly.

Instead of relying on a single-value anomaly metric $a$ a consideration of the reconstruction error vector $\mathbf{e}$ as an metric to flag anomalies may be of interest for future work and will be considered \cite{Hochreiter.1997}.

\section{Model Evaluation}
This Chapter covers the testing and evaluation of the developed anomaly detection models. The models are applied on mission telemetry from Sol 3815 to Sol 3934 (unseen to the models) spanning ~850 m of traversal, simulating a tactical deployment of CAIDDA. We meticulously analyzed these flagged datapoints by cross-referencing them with the rover's flight telemetry, utilizing both telemetry plots and our simulation tool RSVP Hyperdrive \cite{wright_hartman_cooper_maxwell_yen_morrison-06}\cite{verma-leger-19}, which emulates the flight telemetry, providing a simulation of the rover's movement. Both telemetry plots and Hyperdrive simulations approach allowed us to discern the actual rover movements and evaluate how the terrain interaction influenced the rover's behavior. By comparing telemetry data and Hyperdrive simulations, we could confidently determine whether the flagged datapoints truly represented abnormal rover behavior or were, in fact, false detections. This rigorous evaluation process significantly bolstered the precision and credibility of our anomaly detection methodology. In the following, both models are evaluated separately. 

\subsection{CAIDDA-Prime}
The application of CAIDDA-Prime on the queried Sol range resulted in the flagging of 6 samples as anomalous. Table \ref{detected_anomalies_caidda_p} presents the anomalies detected during this specific timeframe, displaying the Martian sol number, reconstruction error $a$, and the top three anomalous features contributing the most to the reconstruction error $a$.

Interestingly, all samples within this query range were flagged with high reconstruction errors for acceleration-related features, particularly those related to vertical accelerations. The analysis using our simulation tool RSVP Hyperdrive revealed that the rover encountered larger rocks, resulting in a suspension response and causing oscillations in vertical acceleration, as shown in Figure \ref{window_size}. Specifically, for 5 out of the 6 flagged samples, the rover's interaction with these rocks led to notable drops, consequently flagging the standard deviation for vertical acceleration as anomalous—logical given the observed oscillations.

While CAIDDA-Prime flagged these instances as anomalous, the mobility team does not perceive the act of the rover dropping off small rocks as an operational risk. However, the ability of CAIDDA-Prime to detect these instances proves valuable to the mobility team. Drops often escape notice during human assessments of a drive. If, for example, the rover were to slip off a rock larger than 20 centimeters, this could pose a risk to the hardware. Given CAIDDA's heightened ability to detect such incidents compared to a human, it significantly reduces the likelihood of potential hardware damage going unnoticed by the downlink team.

Notably, only one of the 6 flagged anomalies did not exhibit any abnormal behavior when analyzed through telemetry and simulation and is therefore marked as a false positive.

\begin{table*}[]
\renewcommand{\arraystretch}{1.3}
\caption{\bf Detected Anomalies by CAIDDA-Prime}
\label{detected_anomalies_caidda_p}
\centering
\begin{tabular}{|l|c|l|l|}
\hline
\textbf{Sol} & \multicolumn{1}{l|}{\textbf{Error a}} & \textbf{Flagged Features}                                     & \textbf{Human Analysis}                          \\ \hline
3858         & 12.1                                               & std(accel{[}Z{]}), max(accel{[}Z{]}), median(accel{[}Z{]})  & Drop off a rock, intense suspension oscillation \\ \hline
3878         & 11.1                                               & median(accel{[}Z{]}), mean(accel{[}Z{]}), min(accel{[}Z{]}) & False detection                                  \\ \hline
3919         & 10.9                                               & std(accel{[}Z{]}), std(accel{[}X{]}), min(accel{[}Z{]})     & Drop off a rock, intense suspension oscillation \\ \hline
3919         & 15.3                                               & std(accel{[}Z{]}), min(accel{[}Z{]}), max(accel{[}Z{]})     & Drop off a rock, intense suspension oscillation \\ \hline
3928         & 10.3                                               & std(accel{[}Z{]}), std(accel{[}X{]}), min(accel{[}Z{]})     & Drop off a rock, intense suspension oscillation \\ \hline
3931         & 32.5                                               & std(accel{[}Z{]}), max(accel{[}Z{]}), mean(accel{[}Z{]})    & Drop off a rock, intense suspension oscillation \\ \hline
\end{tabular}
\end{table*}

\subsection{CAIDDA-Refined}
Similar to the previous section, Table \ref{detected_anomalies_caidda_r} depicts detected anomalies during the specified time frame, showing the Martian sol number, reconstruction error, key anomalous features, and a human analysis of the events during that period. Analyzing the flagged data points, we categorized the anomalies into 5 main categories: Wheelies, Mid-Traverse Startup Current (MTSC), Extensive Slip, Intense Terrain Interaction, and Unknown. Table \ref{detected_anomalies_model_reduced} provides a comprehensive overview of the detected anomalies and their respective classification into specific categories.

Out of the samples, 36.4\% were identified as anomalies, later determined to be bogie suspension wheelies upon human review. Although these detected wheelies were deemed non-harmful to the hardware, the authors were pleasantly surprised by the model's ability to identify them. This detection capability holds promise for potential future scenarios where prolonged friction in the passive bogie joint could cause extended wheel lift, a behavior the model could potentially capture. Detecting wheelies is typically challenging for downlink operators as it requires close examination of hyperdrive drive visualization. CAIDDA's capability to flag samples with wheelies occurring provides an extra layer of safeguarding for the rover.

The MTSC category includes anomalies initiated when the actuator undergoes a planned mid-traverse calibration, a routine event lasting a few seconds, typically during longer drives. This calibration temporarily boosts the current, causing CAIDDA to flag the samples. The model accurately identifies the unusual peak in actuator current, as anticipated. In total 27.3 \% of the samples were assigned to this category by the authors. It is crucial to emphasize that this category does not endanger the rover. Nonetheless, it effectively showcases the model's ability to spot and emphasize unusual patterns. Additionally, it is important to highlight that all MTSC incidents for drive actuators during the query period were accurately flagged by CAIDDA.

\begin{figure}[h]
\centering
\includegraphics[width=3.3in]{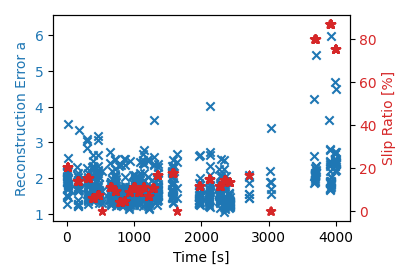}
\caption{\bf{Reconstruction Error a and Slip Ratio for Sol 3839}}
\label{slip_sol3839}
\end{figure}

A significant portion (22.7\%) of the flagged samples were from drive steps characterized by a notably high slip ratio. Out of the five flagged anomalies during drive steps with high slip, four occurred on Sol 3839. To explore the potential connection further, we generated Figure \ref{slip_sol3839} showcasing the reconstruction error ($a$) and rover slip ratio over time for Sol 3839. The plot displays the reconstruction error ($a$) for all drive samples of that Sol on the left axis, alongside the slip ratio of the drive steps on the right axis. It is important to highlight that the rover calculates the mean slip ratio for each entire drive step. Upon observing the plot, it becomes evident that there is a general increase in error for the last three drive steps. In that Sol, the three drive steps with the highest slip exhibit the most noticeable errors. While it cannot be asserted with absolute certainty that the increase in error ($a$) is solely due to extensive slipping, the notable correlation strongly suggests a connection.

One of the flagged samples, upon thorough review by the authors, was associated with a significant interaction with the terrain (Intense Terrain Interaction). The left middle (LM) wheel loaded against a rock and attempted to traverse it. During this maneuver, the current of the drive actuator of that wheel surged and the rover started to shake intensely. The rover was unable to surmount the rock and instead moved laterally, resulting in substantial suspension movement. Importantly, there was no longitudinal drive progress during this particular step. From an operational standpoint, the mobility team assessed that this drive step did not pose any safety risk to the rover. However, it remains an intriguing and noteworthy drive step.

Upon review, it was determined that two samples flagged as anomalies by CAIDDA-Refined did not demonstrate any abnormal behavior. Consequently, they have been marked as false positives.

\begin{table}[h]
\renewcommand{\arraystretch}{1.3}
\caption{\bf Overview of Detected Anomalies by CAIDDA-Refined}
\label{detected_anomalies_model_reduced}
\centering
\begin{tabular}{|l|c|c|}
\hline
\textbf{Anomaly Type}        & \multicolumn{1}{l|}{\textbf{Amount}} & \multicolumn{1}{l|}{\textbf{Percentage}} \\ \hline
Wheelie                      & 8                                    & 36.4 \%                                  \\
Mid-Traverse Startup Current & 6                                    & 27.3 \%                                  \\
Extensive Slip               & 5                                    & 22.7 \%                                  \\
Unknown                      & 2                                    & 9.1 \%                                  \\
Intense Terrain Interaction                      & 1                                    & 4.5 \%                                  \\ \hline
\textbf{Total}               & \textbf{22}                          & \textbf{100 \%}                          \\ \hline
\end{tabular}
\end{table}

\begin{table*}[]
\renewcommand{\arraystretch}{1.3}
\caption{\bf Detected Anomalies by CAIDDA-Refined}
\label{detected_anomalies_caidda_r}
\centering
\begin{tabular}{|c|c|l|l|}
\hline
\multicolumn{1}{|l|}{\textbf{Sol}} & \multicolumn{1}{l|}{\textbf{Error a}} & \textbf{Flagged Features}                              & \textbf{Human Analysis}          \\ \hline
3839                               & 5.45                                               & kurt(P{[}RF{]}), kurt(C{[}RF{]}), min(Rate{[}RR{]}), & 86 \% wheel slip during drive \\ \hline
3839                               & 5.98                                               & kurt(PD{[}LF{]}), kurt(PD{[}RR{]}), kurt(CD{[}LF{]}) & 89 \% wheel slip during drive                                 \\ \hline
3839                               & 4.68                                               & std(V{[}LR{]}), std(CD{[}LR{]}), median(C{[}LR{]})   & 81 \% wheel slip during drive                                 \\ \hline
3839                               & 4.50                                               & min(C{[}LR{]}), kurt(P{[}LF{]})                      & 81 \% wheel slip during drive                                 \\ \hline
3839                               & 4.62                                               & median(RATE{[}LM{]}), kurt(C{[}LM{]})                & 79 \% wheel slip during drive                                 \\ \hline
3871                               & 6.14                                               & min(CD{[}LR{]}), min(PD{[}LR{]}), std(PD{[}LR{]})    & Wheelie Build-Up and Recovery LM                              \\ \hline
3871                               & 4.56                                               & kurt(P{[}LM{]}), skew(CD{[}LM{]}), skew(PD{[}LM{]})  & Wheelie Recovery LR                                           \\ \hline
3873                               & 5.28                                               & kurt(C{[}RF{]}), kurt(P{[}RF{]})                     & Wheelie Build-Up LR                                           \\ \hline
3880                               & 5.40                                               & skew(CD{[}RF{]})                                     & Wheelie Build-Up LM                                           \\ \hline
3894                               & 5.49                                               & min(PD{[}RR{]}), min(CD{[}RR{]})                     & Wheelie Recovery RM                                           \\ \hline
3897                               & 5.11                                               & max(CD{[}LR{]}), max(C{[}LR{]})                      & MTSC LR                                                       \\ \hline
3899                               & 5.00                                               & median(PD{[}LR{]}), median(CD{[}LR{]})               & Wheelie Build-Up LM                                           \\ \hline
3899                               & 4.66                                               & min(CD{[}RM{]}), min(PD{[}RM{]})                     & Unknown                                                       \\ \hline
3901                               & 5.39                                               & max(CD{[}RR{]}), max(C{[}RR{]})                      & MTSC RR                                                       \\ \hline
3901                               & 5.75                                               & max(CD{[}LF{]}), max(C{[}LF{]})                      & MTSC LF                                                       \\ \hline
3910                               & 4.85                                               & max(CD{[}LR{]}), max(C{[}LR{]}                       & MTSC LR                                                       \\ \hline
3917                               & 6.29                                               & max(CD{[}RR{]}), max(C{[}RR{]})                      & MTSC RR                                                       \\ \hline
3917                               & 6.37                                               & max(C{[}RR{]})                                       & Unknown                                                       \\ \hline
3919                               & 4.88                                               & min(PD{[}LR{]}), min(C{[}LR{]})                      & Wheelie Recovery LR                                           \\ \hline
3919                               & 4.51                                               & max(PD{[}LM{]}), std(PD{[}LM{]})                     & Intense Terrain Interaction                                   \\ \hline
3931                               & 5.33                                               & min(P{[}RM{]}), min(C{[}LR{]})                       & Wheelie Build-Up RM                                           \\ \hline
3931                               & 8.04                                               & max(CD{[}LM{]}), max(C{[}LM{]})                      & MTSC LM                                                       \\ \hline
\end{tabular}
\end{table*}

\section{Conclusion and Future Work}
In this study, we have presented our comprehensive methodology, providing insights into our data sources and the incorporation of mobility signals. Our preprocessing approach involved the computation of additional telemetry signals, enhancing the depth of our analysis. We introduced and evaluated two autoencoder-based anomaly detection models: CAIDDA-Prime and CAIDDA-Refined. CAIDDA-Prime, leveraging all input features, including acceleration-related signals, demonstrated heightened sensitivity to anomalies associated with acceleration. On the other hand, CAIDDA-Refined, using a filtered input excluding acceleration signals, showcased a broader capacity to detect anomalies related to actuator telemetry.

Our evaluation, encompassing unseen data over an extensive rover traversal of 850 meters, revealed the efficacy of our models. The flagged samples, identified using a threshold of the 99.9th percentile, were meticulously analyzed. Notably, CAIDDA-Prime detected anomalies predominantly linked to acceleration-related features, exposing intense terrain interaction during hyperdrive simulation review. While these anomalies did not present an immediate operational risk, their detection holds significant value for the mobility team. Drops and intense terrain interaction often escape human assessment, making our detection capability critical in averting potential hardware damage, especially in scenarios where the rover could descend from larger rocks. On the other hand, CAIDDA-Refined identified anomalies in actuator telemetry, including wheelies, extensive slipping, and mid-traverse startup currents. It is important to note that these detected anomalies did not present any risk to the rover's health and safety. However, they illustrated the model's potential to flag samples in the event of such occurrences, providing a proactive warning to operators. Further testing of these models using the Curiosity Vehicle System Testbed (VSTB), a Mars rover duplicate situated at our research facilities in the Jet Propulsion Laboratory, is required to verify the model's capability to detect severe anomalies.

The work presented in this paper underscores the critical role of feature selection in effective anomaly detection. When employing all input features with CAIDDA-Prime, we observed a model bias towards acceleration-related anomalies. However, when we filtered the input features, as demonstrated with CAIDDA-Refined, the model detected more subtle anomalies that were initially overlooked by CAIDDA-Prime. It suggests the potential necessity for a weight vector that assigns higher priority to the reconstruction error of specific features over others. This proposition warrants a thorough analysis of the data, which would be a topic of future investigation by the authors. 

The autoencoder architecture underwent manual optimization. As mentioned earlier, further optimization could be achieved through a hyperparameter sweep. Additionally, experimenting with varying bottleneck sizes while aiming for an acceptable loss is a viable approach. Alternatively, exploring a CNN \cite{LeCun.2015} or LSTM \cite{Hochreiter.1997} based approach could uncover patterns across timesteps and potentially identify nuanced anomalies that might be missed due to the limitation of extracting only 7 statistical features. However, it is worth noting that while a CNN or LSTM based approach may offer enhanced anomaly detection capabilities, it may lack the same level of model interpretability as a feature-extraction-based approach.

Finally, the analysis and evaluation of flagged anomalies underscored CAIDDA's potential to detect anomalies revealed through mobility telemetry. The authors intend to integrate CAIDDA-Prime and CAIDDA-Refined into the tactical downlink process for ongoing validation of the models' anomaly detection capabilities throughout the mission. The anomalies detected within the queried period demonstrate the model's aptitude in detecting nuanced deviations in telemetry and highlight its potential to identify more significant deviations that could endanger the rover and the mission.


\acknowledgments
The research described in this paper was performed at the Jet Propulsion Laboratory, California Institute of Technology, under contract with the National Aeronautics and Space Administration(80NM0018D0004). The authors would like to thank the Mars Science Laboratory Program for supporting this research.

\bibliographystyle{IEEEtran}
\bibliography {biblio}

\thebiography
\begin{biographywithpic}
{Mielad Sabzehi}{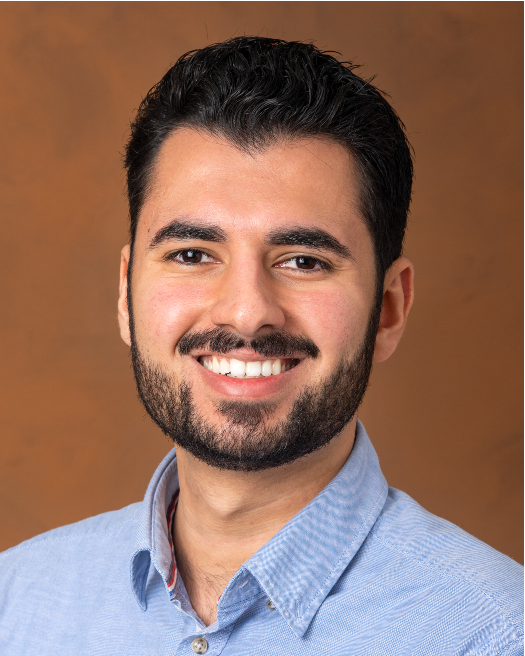}
is a Robotics Technologist in the Robotic Mobility Group at JPL. He is a member of the Mobility/Mechanisms subsystem team for MSL and Robotic Arm subsystem team for M2020. His primary areas of research include machine learning-based mobility optimization and anomaly detection. He received his MSc degree in Mechanical Engineering from RWTH Aachen University in 2023, where he specialized in Computer Science and Aerospace Engineering. 
\end{biographywithpic} 

\begin{biographywithpic}
{Peter Rollins}{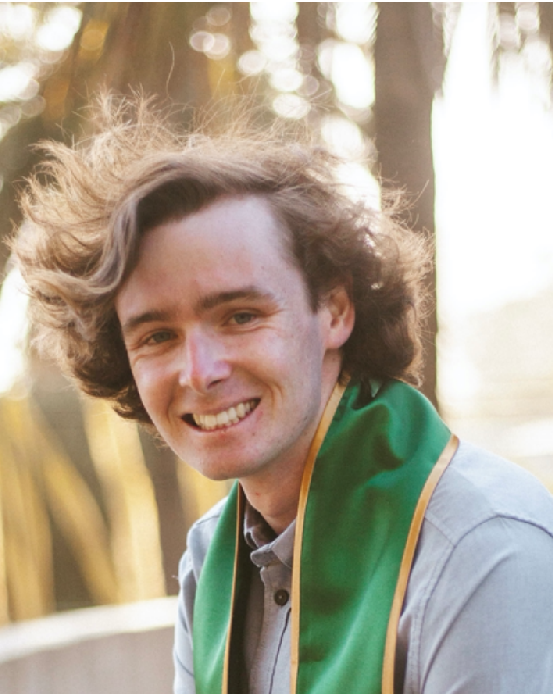}
is a Systems Engineer in the Engineering Operations for Surface Missions group at JPL. He is a member of the MSL Operations Team, where he leads the Mobility and Mechanisms Subsystem in both tactical downlink and long-term strategic tasks. He received his B.S. in Aerospace Engineering from California Polytechnic State University, San Luis Obispo.s

\end{biographywithpic}

\end{document}